\documentclass[runningheads]{llncs}
\usepackage{graphicx}
\usepackage{comment}
\usepackage{amsmath,amssymb} 
\usepackage{color}
\usepackage{cite}
\usepackage{enumitem}
\usepackage[marginal]{footmisc}

\begin{document}
\pagestyle{headings}
\mainmatter
\def\ECCVSubNumber{3293}  

\title{Graph Edit Distance Reward: Learning to Edit Scene Graph} 

\titlerunning{Graph Edit Distance reward: Learning to edit scene graph}

\author{Lichang Chen\inst{1, 2^\dagger} \and
Guosheng Lin\inst{1^\ddagger} \and
Shijie Wang\inst{1, 3} \and Qingyao Wu\inst{4, 5} }
\authorrunning{L. Chen, G. Lin et al. }
\institute{Nanyang Technological University \and Zhejiang University  \and Huazhong University of Science and Technology \\ \and
School of Software Engineering, South China University of Technology \\ \and Key Laboratory of Big Data and Intelligent Robot, Ministry of Education \\
\email{bobchen@zju.edu.cn, gslin@ntu.edu.sg}}
\maketitle

\begin{abstract}
\label{section: abstract}
\par Scene Graph, as a vital tool to bridge the gap between language domain and image domain, has been widely adopted in the cross-modality task like VQA. In this paper, we propose a new method to edit the scene graph according to the user instructions, which has never been explored. To be specific, in order to learn editing scene graphs as the semantics given by texts, we propose a Graph Edit Distance Reward, which is based on the Policy Gradient and Graph Matching algorithm, to optimize neural symbolic model. In the context of text-editing image retrieval, we validate the effectiveness of our method in CSS and CRIR dataset. Besides, CRIR is a new synthetic dataset generated by us, which we will publish soon for future use.   

\keywords{ Scene Graph Editing, Policy Gradient, Graph Matching}
\end{abstract}

\footnote{\noindent $^\dagger$ This work was done when L. Chen was an intern student in NTU. \\$^\ddagger$ G. Lin is the corresponding author of this paper.}
\begin{figure}[t]
\begin{center}
\includegraphics[width=0.75\linewidth]{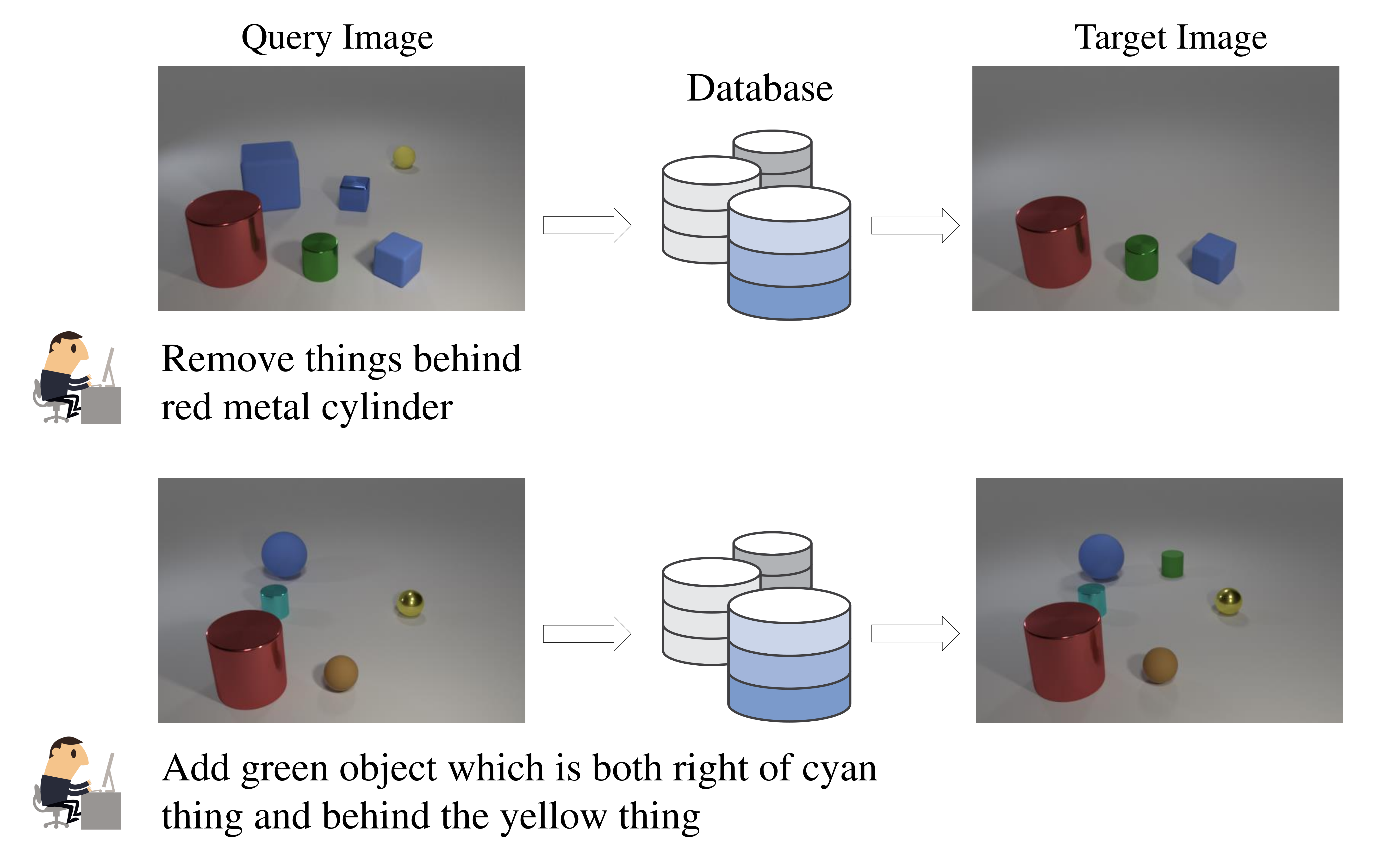}
\end{center}
   \caption{Examples about cross-modal image retrieval. Given a query image and a query text describing the modification to the query image, we expect to edit the scene graph of the query image then retrieve a target image from database.}
\label{fig: Introduction}
\end{figure}
\section{Introduction}
\label{section: introduction}

\par Nowadays, in our daily life, more and more people have grown accustomed to typing in texts to obtain images from search engine. However, if they are not satisfied with searching results or they would like to modify the previous results, the only thing they can do is altering the language, then searching again. If users can provide their instructions about how to edit the searching results, it will be more than convenient for them to retrieve satisfied images. Figure 1 shows the simulation of such a scenario in the synthetic dataset. Vo \textit{et al.}\cite{vo2019composing} firstly propose this text-editing image retrieval task, where the input query is composed by an input image $I_{input}$ and an input text $T_{input}$, describing the desired modifications of $I_{input}$. Here, $T_{input}$ can be viewed as user instructions and $I_{input}$ can be viewed as the unsatisfactory image. We believe it is a superb platform for simulating the cross-modal task in real life.  

\par Teaching machines to comprehend interaction between language and vision is an essential step for solving the Text+Image problem. There are some visual question answering(VQA\cite{antol2015vqa, johnson2017clevr}) systems\cite{johnson2017inferring, yi2018neural} succeeding in crossing modalities by reasoning over Scene Graphs\cite{johnson2015image} following the semantics given by the texts, which proves Scene Graphs can bridge the gap between vision and language. Some image captioning models\cite{yang2019auto, luo2018discriminability} also adopted Scene Graphs and achieves good results, demonstrating that Scene Graphs are a fantastic representation of image scenes. We represent our scene by Scene Graph owing to the advantages of Scene Graphs. In this way, we transform the image editing problem into Scene Graph editing problem.

\begin{figure*}[t]
\begin{center}
\includegraphics[width=1\linewidth]{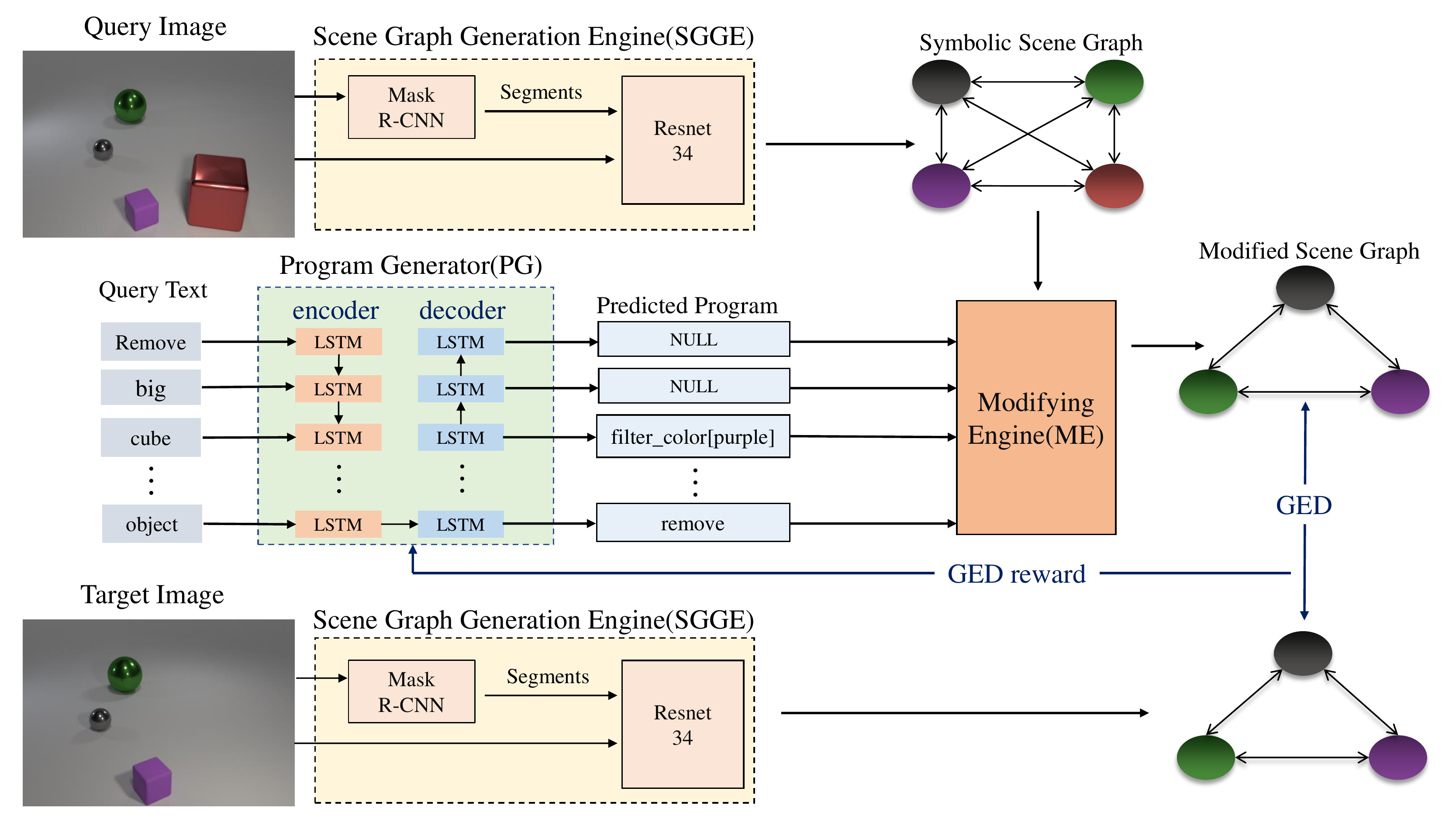}
\end{center}
\caption{Model Overview. The intact query text is ``remove big cube to the right of purple object''. The intact Predicted Program is ``remove'', ``filter\_size[large]'', ``filter\_shape[cube]'', ``relate[right]'', ``filter\_color[purple]''. Because the query text is longer than the predicted program, our Program Generator will pad ``NULL'' program to fix the length inconsistency.}
\label{fig: model_overview}
\end{figure*}

\par Due to the booming of computation resources, ``old school'' symbolic models lose its position from the competition with numeric model \cite{minsky1991logical}, which need less human intervening. However, it is undeniable that symbolism has one secret weapon, neat representation, which the high dimensional numeric does not possess. Recently, utilizing the advantage of symbolism, Yi \textit{et al.} \cite{yi2018neural} propose applying neural symbolic model on CLEVR \cite{johnson2017clevr} with fully transparent reasoning. In details, firstly they use a scene parser to derender the scenes from images. Then they train a seq2seq program generator to predict the latent program from a question. Finally, their program executor will execute these predicted programs over symbolic scene representations to produce answer. Motivated by their model's neat representations and fully explainable characteristics, we employ it to parse the user instructions. We design a Scene Graph Generation Engine to parse each image into a symbolic scene graph, whose nodes represent the objects while edges represent the relation between nodes. We also design a module to edit the scene graph. After that, the retrieval problem can be converted to a graph matching problem \cite{conte2004thirty}, which has low combinatorial space due to the symbolic representation.

\par We notice that REINFORCE \cite{williams1992simple} algorithm applied to finetune the Neural Symbolic Model\cite{yi2018neural} is the most vital part in achieving marvelous results. Following Johnson \textit{et al.} \cite{johnson2017inferring}, Yi \textit{et al.} apply $ 0, 1 $ reward to finetune their program generator. This reward, however, is quite coarse if applying to our task. To be specific, only when exact matching \cite{conte2004thirty} happens, which means all the nodes and all the edges in two generated symbolic scene graphs(one 
is generated from query image, another is generated from target image) matching, it is set to 1. To refine it, we propose applying Graph Edit Distance \cite{sanfeliu1983distance} based reward, which can optimize our model better.

\par In our experiment, we discover the CSS dataset \cite{vo2019composing} is simple because it has plenty of exact location information. To validate our model's efficacy further, we follow CLEVR's \cite{johnson2017clevr} generation process and generate our own Complex Reasoning Image Retrieval, called CRIR, which contains multi-hop queries thus the scene graphs are hard to be accurately edited. Moreover, our dataset breaks the constraints of exact location so it is more close to the real world scenes.

\par To summarize, our contributions are:
\begin{itemize}
    \item We propose a GED reward to explore editing the scene graph with user instruction in the context of text-editing image retrieval. We refine the previous neural symbolic model aiming at VQA and make it suitable for our cross-modal image retrieval task. To the best of our knowledge, we are the first to propose editing the scene graph. 
    
    \item We propose a new dataset, CRIR, which contains abundant complex queries with full annotations of images. The dataset breaks the constraints of exact location in CSS \cite{vo2019composing}, which can be used to validate the generalization ability of our model and simulate the real-world image retrieval scene better.
    
    \item Based on the policy gradient algorithm\cite{williams1992simple}, we propose Graph Edit Distance(GED) reward to finetune our model as well as apply GED as retrieval metric, making it learn to edit scene graphs better. In the context of text-editing image retrieval task, we validate the efficacy of our model. Notably, we achieve new state-of-the-art performance on both CRIR and CSS\cite{vo2019composing}, surpassing the previous methods by large margins. 
\end{itemize}
\begin{figure*}[t]
\begin{center}
\includegraphics[width=1\linewidth]{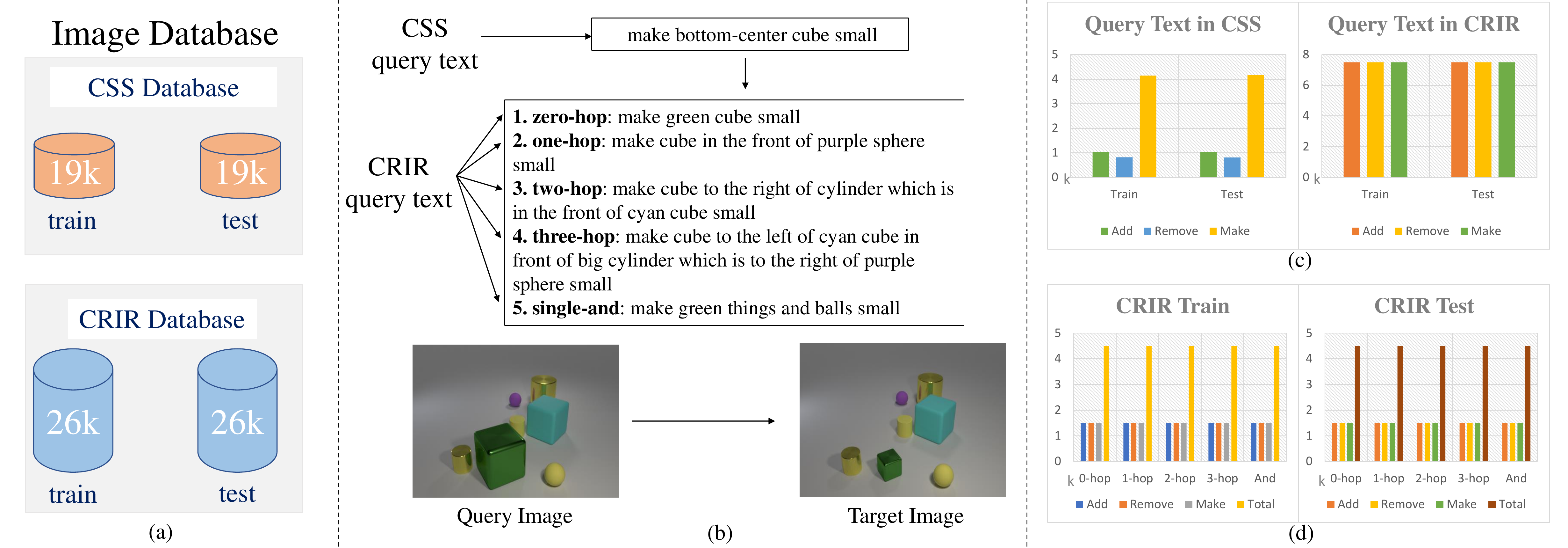}
\end{center}
   \caption{(a) Image database scale -- CRIR vs. CSS. (b) CSS only has zero-hop query. CRIR has 5 different types of queries. (c) Query Text Scale -- CRIR vs. CSS. In CSS dataset, both training set and testing set contain 6k queries, while in CRIR dataset, both training set and testing set have 4.5k queries in each type of templates, totally 22.5k queries. (d) Query text type details. CRIR eliminates the dataset bias, each template generating the same number of queries.}
\label{fig: CRIR Dataset}
\end{figure*}

\section{Related Work}
\label{section: related work}

\par \textbf{Image Retrieval.} Conventional image retrieval \cite{lin2015deep} is a task to retrieve the most similar image to the query image from database, which is also viewed as image to image matching problem. In recent year, with much stronger image feature extractor\cite{He_2016_CVPR, simonyan2014very}, conventional image retrieval has also made a huge progress\cite{liu2016deep, babenko2015aggregating, gordo2016deep} and has been extended to different real-world scenes like face retrieval. \cite{schroff2015facenet, liu2016deepfashion}.

\par Johnson \textit{et al.} \cite{johnson2015image} first proposed cross-modality image retrieval. The task can be defined as given a text $T_{input}$, the model should retrieve an image $I_{target}$ which is the most relevant to the $T_{input}$. They also give the definition of scene graph and use it to represent the content of the scene, which builds a bridge between image domain and text domain. What's more, cross-modality image retrieval is also extended to real-world scenes such as recipe to food image retrieval where the input query is recipe, food image to recipe retrieval, using deep metric learning method\cite{Wang_2019_CVPR}.

\noindent \textbf{Visual Question Answering.} Visual Question Answering \cite{antol2015vqa} requires model to answer question given by text based on the input image. Solving this problem is a vital step to cross the modalities between language and vision.

\par Recently, Johnson \textit{et al.} \cite{johnson2017clevr} utilize Blender to render images and they obtain a synthetic, virtual, and diagnostic dataset named CLEVR with full annotations of objects, such as 3D coordinates, size, shape, color, material, etc. The low cost of rendering virtual datasets allows for the appearance of CLEVR based datasets. For instance, Liu \textit{et al.}\cite{liu2019clevr} refine the CLEVR program generation engine and propose CLEVR-REF+ for referring expression task. They also modify the rendering engine which generates images with segments and bounding box information. There are also some real-world VQA datasets like VQA v1.0, v2.0 \cite{antol2015vqa} and GQA \cite{hudson2019gqa}, etc. Though the annotations cost is high, they provide a more realistic platform for validating the methods.

\noindent \textbf{Neural Module Network.} NMN is firstly proposed by Andreas \textit{et al.} \cite{andreas2016neural}. They parsed the natural language sentences into programs and applied it to instantiate different neural modules, then executing them to obtain the answer. Following their work, Johnson \textit{et al.} \cite{johnson2017inferring} and Hu \textit{et al.} \cite{hu2017learning} extend the Neural Module Network to the CLEVR dataset. But their neural modules are numeric one with executing programs and reasoning on high dimensional features. Yi \textit{et al.} \cite{yi2018neural} propose symbolic module network to execute programs over symbolic scenes generated by scene parser achieving new state-of-the-art result in CLEVR, 99.9\%. Due to the efficacy of their model, we adapt them to text-editing image retrieval task.

\begin{figure*}[t]
\begin{center}
\includegraphics[width=1\linewidth]{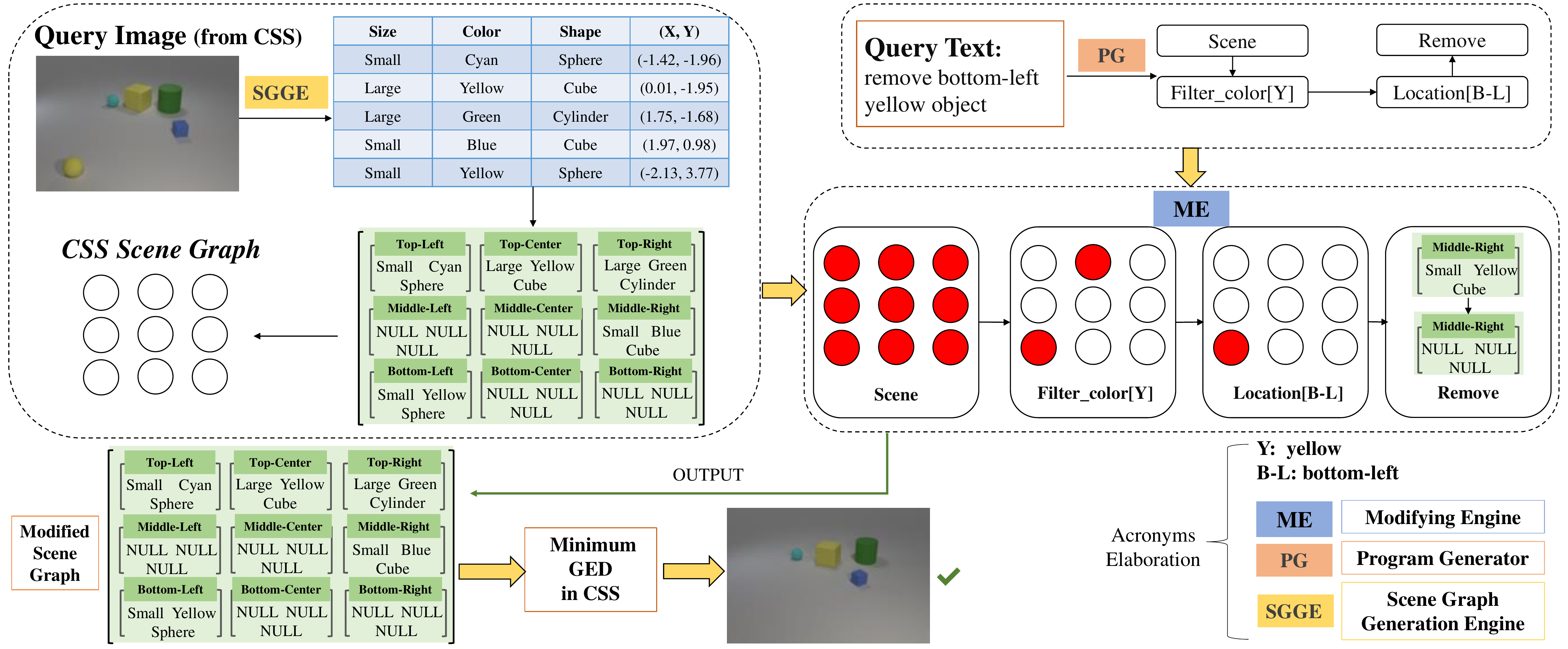}
\end{center}
   \caption{A scene graph editing case in CSS dataset. CSS Scene Graph does not have any edges. So Modifying Engine just reason over nodes. In final {\tt remove} program, we keep structure of the scene graph by just substituting the attribute values in operation. Red nodes in ME means attended nodes. Minimum GED means the minimum Graph Edit Distance between Modified Scene Graph and other Scene Graphs in CSS.}
\label{fig: CSS reasoning}
\end{figure*}

\section{Method}
\label{section: method}
\par In section \ref{Our model}, we introduce our model. In section \ref{graph edit distance}, we give the definition of Graph Edit Distance. In section \ref{GED + policy gradient}, we introduce policy gradient with new method we proposed to estimate the gradient in this task.

\subsection{Our Model}
\label{Our model}
\par  Figure \ref{fig: model_overview} shows the overview of our model. Our model is adapted from neural symbolic reasoning model in \cite{yi2018neural}. The first step to train our model is supervised training separately on both Scene Graph Generation Engine (SGGE) and Program Generator (PG) with Images and small number of ground truth latent programs(We provide the training details in Section \ref{section: experiment on css} and \ref{section: experiment on CRIR} ). In finetuning, the input of our model contains $T_{input}$ and $I_{input}$. From SGGE, We obtain the symbolic scene graph of the $I_{input}$ while the latent program can be parsed by PG. The output of PG will be put into Modifying Engine (ME) with the generated scene graph. Then ME will execute programs sequentially and output the modified scene graph. Finally, the modified scene graph is compared with the target scene graph, which is generated from target image by the same SGGE, in Graph Edit Distance metric. After the computation of GED, the GED based reward can be backpropagated to finetune the PG.
\par The details about three main components, Scene Graph Generation Engine, Program Generator, Modifying Engine show as follows: 

\par \textbf{Scene Graph Generation Engine.} We use SGGE to obtain the 3D coordinates with attribute values, including shape, size, color, material, for each objects in the image. In detail, Mask R-CNN is applied to generate segment proposals of each object in the image. Then we input the segment of each object with original image, resized to 224 by 224, into a Resnet-34 \cite{he2016deep} to predict 3D coordinates.

\par \textbf{Program Generator.} The program generator follows the language model proposed by \cite{yi2018neural} which is improved version of \cite{johnson2017inferring} with attention. It is a seq2seq model. We set the dimension of input word vector to 300, 2 hidden layers in both encoder and decoder to 256 dimension. As the Figure \ref{fig: model_overview} shows, we return a reward from the final part to finetune our program generator. Because the Modifying Engine (ME) execution part is symbolic, the gradient stream will be cut down there, which means the model is undifferentiable when entering ME. Therefore, we exploit small number of ground truth programs to pretrain our PG, then using GED reward to finetune it. Because training and finetuning details are different in different datasets, we will elaborate them in Experiment (Section \ref{section-5}).
\par \textbf{Modifying Engine.} The modifying engine will execute programs predicted by program generator over symbolic scene graphs. Thanks to the neat representation of symbolic representation, the reasoning and modifying operations over symbolic scene graphs are fully transparent and make our image retrieval process explainable. We adopt different modifying engines in two experiments. The details about 2 modifying engines show in Section \ref{section: experiment on css} and \ref{section: experiment on CRIR}. We implement our Modifying Engine as a set of modules in Python. Each program has one counterpart module in ME. The programs will first be converted to the corresponding modules and then be executed in Modifying Engine sequentially.

\subsection{Graph Edit Distance.}
\label{graph edit distance}
\par Graph Edit Distance(GED) is a graph matching \cite{conte2004thirty} approach first proposed by Sanfeliu \textit{et al.} \cite{sanfeliu1983distance}. The concept of GED is finding the optimal set of edit operation which can transform Graph $G_1$ into Graph $G_2$.\\
\textbf{Definition1.} (Graph $G$)\\
A graph $G$ can be represented by a 3-tuple $\left(V, \alpha, \beta\right)$, such that: V is a set of nodes. $\alpha: V \rightarrow L$ is the node labeling function.  $\beta: V \times V \rightarrow L$ is the edge labeling function. \\
\textbf{Definition2.} (Graph Isomorphism)\\
A graph isomorphism between $G_1 = \left(V_{1}, \alpha_{1}, \beta_{1}\right)$ and $G_2 = \left(V_{2}, \alpha_{2}, \beta_{2}\right)$ is a bijective mapping $f: V_{1} \rightarrow V_{2}$ such that $\alpha_{1}(x)=\alpha_{2}(f(x))$ for all $x \in V_{1}$ and $\beta_{1}((x, y))=\beta_{2}((f(x), f(y)))$ for all $(x, y) \in V_{1} \times V_{1}$.\\
\textbf{Definition3.} (Common Subgraph) \\
Let $G_{1}^{\prime} \subseteq G_{1}$ and $G_{2}^{\prime} \subseteq G_{2}$, if there exists a graph isomorphism between $G_{1}^{\prime}$ and $G_{2}^{\prime}$, both $G_{1}^{\prime}$ and $G_{2}^{\prime}$ will be called a common subgraph of $G_1$ and $G_2$. Moreover, a graph G is called a maximum common subgraph of $G_1$ and $G_2$ if $G$ is a common subgraph of $G_1$ and $G_2$ and there exists no other common subgraph of $G_1$ and $G_2$ that has more nodes than $G$.\\
\textbf{Definition4.} (error-correcting graph matching) \\
An error-correcting graph matching from $G_1$ to $G_2$ is a bijective function $f: \hat{V}_{1} \rightarrow \hat{V}_{2}$ where $\hat{V}_{1} \subseteq V_1$ and $\hat{V}_{2} \subseteq V_{2}$.\\
\textbf{Definition5.} (Graph Edit Distance) \\
Let $G_{1}=(V_{1}, \alpha_{1},\beta_{1})$, $G_{2}=(V_{2}, \alpha_{2}, \beta_{2})$ be two graphs, the GED between these 2 graphs is defined as
\begin{equation}
    GED\left(G_1, G_2\right)=\min _{e_{1}, \cdots, e_{k} \in \gamma\left(f\right)} \sum_{i=1}^{k} c\left(e_{i}\right),
\label{GED}
\end{equation}
where $f$ is an error-correcting graph matching $f: \hat{V}_{1} \rightarrow \hat{V}_{2}$ from Graph $G_1$ to Graph $G_2$ and $c$ denotes the cost function measuring the strength $c(e_i)$ of an edit operation $e_i$ and $\gamma(f)$ denotes the set of edit paths transforming $G_1$ into $G_2$. Insertions, deletions, and substitutions of both edges and nodes are 4 types of edit operations allowed. Thus, the right part of equation \ref{GED} can be denoted as \cite{bunke1999error}
\begin{equation}
\begin{aligned}
\min_{e_{1}, \cdots, e_{k} \in \gamma(G_{1}, G_{2})}\sum_{i=1}^{k} c\left(e_{i}\right)= \min\Big(\sum_{x \in V_1 - \hat{V_1}}{c_{nd}(x)} + \\\sum_{x \in V_2 - \hat{V_2}}{c_{ni}(x)} + \sum_{x \in \hat{V_1}}{c_{ns}(x)} + \sum_{e \in \hat{E_1}}{c_{es}(e)}\Big),
\end{aligned}
\label{ecgm}
\end{equation}
where $c_{nd}(x)$ is the cost of deleting a node $x \in V_1 - \hat{V_1}$ from $G_1$. $c_{ni}(x)$ is the cost of inserting a node $x \in V_2 - \hat{V_2}$ in $G_2$, $c_{ns}(x)$ is the cost of inserting a node $x \in \hat{V_1}$ by $f(x) \in \hat{V_2}$, and $c_{es}(e)$ is the cost substituting a node $x \in \hat{V_1}$ by $f(x) \in \hat{V_2}$ and $c_{es}(e)$ is the cost of substituting an egde $e = (x, y) \in \hat{V_1} \times V_1 $ by $e' = (f(x), f(y)) \in V_2 \times \hat{V_2}$. The cost of 4 operations are hyperparameter in our model. In experiment part\ref{section-5}, we will show the best hyperparameters in different datasets.
\par We follow the widely used quick GED computation algorithm proposed by Riesen \textit{et al.} \cite{riesen2007speeding} which is referred to as A*GED to compute GED in our paper.

\subsection{Policy Gradient + GED reward}
\label{GED + policy gradient}
In Reinforcement Learning, unlike the image classification and segmentation, the problems are usually undifferentiable with high-dimensional discrete states and learning (state, action) pair is very hard. However, the policy, such as make a robot move left or kick a ball, is easier to learn. Thus, we can formally define a class of parametrized policies as $\Pi=\left\{\pi_{\theta}, \theta \in \mathbb{R}^{m}\right\}$. And for each policy, the value is defined as equation \ref{policy}, where $r_t$ is the reward in the time step t, $\gamma^t$ is the attenuation coefficient in the time step t.
\begin{equation}
    J(\theta)=\mathbb{E}\left[\sum_{t \geq 0} \gamma^{t} r_{t} | \pi_{\theta}\right]
\label{policy}
\end{equation}
Our goal is to find the optimal policy $\theta^{*}=\arg \max _{\theta} J(\theta)$. And one of simple algorithms to solve this problem is REINFORCE \cite{williams1992simple} which applies gradient ascent to optimize the parameter $\theta$. The expected future reward can be written as
\begin{equation}
\label{expected reward}
\begin{aligned} J(\theta) &=\mathbb{E}_{\tau \sim p(\tau ; \theta)}[r(\tau)] \\ &=\int_{\tau} r(\tau) p(\tau ; \theta) \mathrm{d} \tau, \end{aligned}
\end{equation}
where the $p(\tau ; \theta)$ is the probability of trajectory $\tau$ in sampling. And by differentiating the equation \ref{expected reward}, we can obtain:
\begin{equation}
\nabla_{\theta} J(\theta)=\int_{\tau} r(\tau) \nabla_{\theta} p(\tau ; \theta) \mathrm{d} \tau.
\label{eq: policy}
\end{equation}
$\nabla_{\theta} p(\tau ; \theta)$ in equation \ref{eq: policy} can be rewritten as
\begin{equation}
\label{t}
    \nabla_{\theta} p(\tau ; \theta)=p(\tau ; \theta) \frac{\nabla_{\theta} p(\tau ; \theta)}{p(\tau ; \theta)}=p(\tau ; \theta) \nabla_{\theta} \log p(\tau ; \theta).
\end{equation}
Finally, we can estimate $J(\theta)$ as equation \ref{t}:
\begin{equation}
\label{REINFORCE algorithm}
    \nabla_{\theta} J(\theta) \approx \sum_{t \geq 0} r(\tau) \nabla_{\theta} \log \pi_{\theta}\left(a_{t} | s_{t}\right),
\end{equation}
where $s_{t}$ is the state in time step $t$.
\par As previously state in Section \ref{section: introduction}, the simple $ 0, 1 $ reward is coarse in our task. Thanks to the graph matching algorithm, we have a better measurement about the similarity of 2 graphs. To measure the reward after a number of programs executing in Modifying Engine better, we propose a new reward (See equation \ref{new reward}) combining Graph Edit Distance with policy gradient algorithm named GED reward. In experiment, we will validate the efficacy of this reward when it is applied to finetune our program generator. 
\begin{equation}
\label{new reward}
    reward = 1 - GED.
\end{equation}

\begin{figure*}[t]
\begin{center}
\includegraphics[width=1\linewidth]{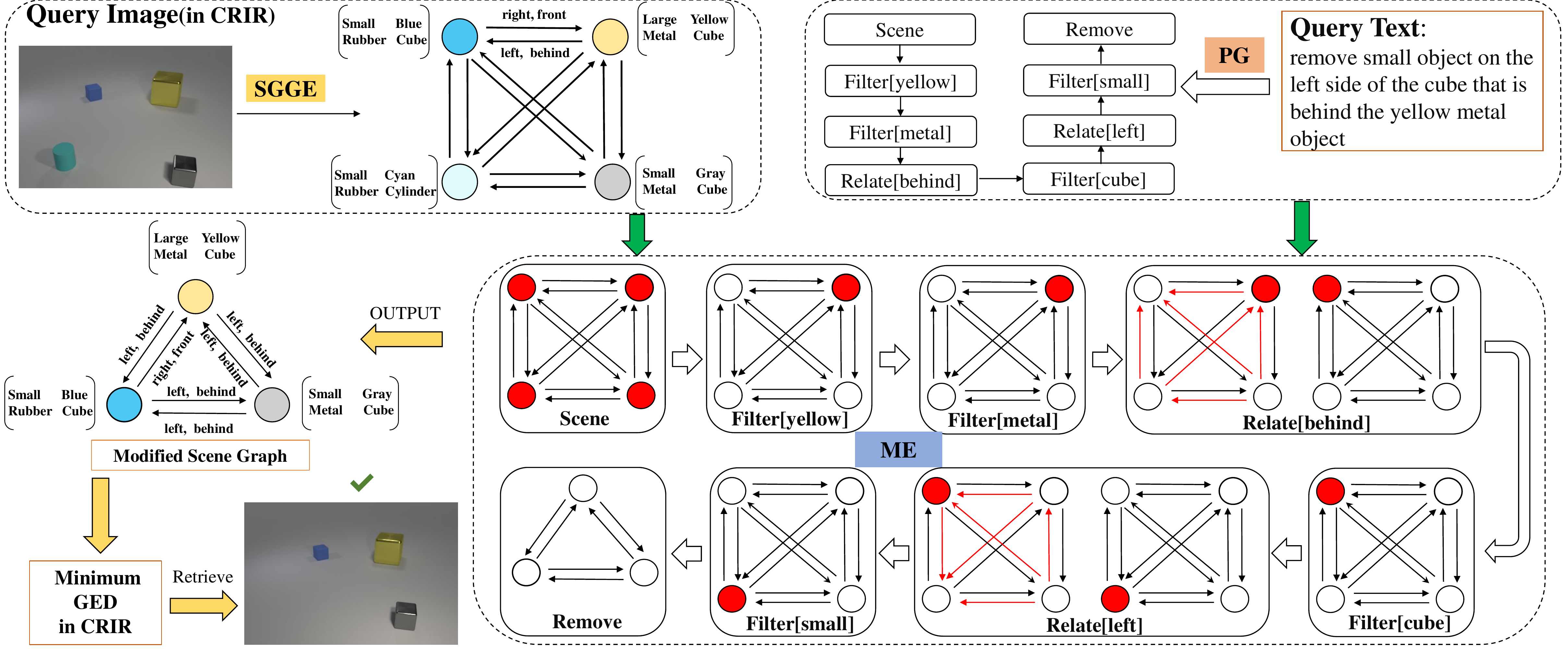}
\end{center}
   \caption{A scene graph editing case in CRIR dataset. ME is Modifying Engine. PG is Program Generator. SGGE is Scene Graph Generation Engine. Symbolic Reasoning over CRIR scene graphs need utilizing edge information, as shows in ``Relate'' program reasoning part. Here, we display the reasoning details of ME. The reasoning process in CRIR is more complex than in CSS.}
\label{fig: CRIR Reasoning}
\end{figure*}

\section{CRIR Dataset}
\label{section: CRIR Dataset}

\par After exploring the Color, Shape, Size(CSS) \cite{vo2019composing} dataset carefully, we find it just adopt one simplest kind of template, zero-hop, in CLEVR's\cite{johnson2017clevr} template universe, which means it does not need any reasoning process before editing the scene graphs. Thus, inheriting the CLEVR template universe, we generate more complex reasoning dataset called CRIR to validate the reasoning and generalization ability of our model. 

\par Our dataset generation can be split into 2 parts, image generation and query text generation, which will be discussed in section \ref{subsection: Image Generation} and section \ref{subsection: Query Generation} respectively.  

\subsection{Image Generation}
\label{subsection: Image Generation}
\par Firstly, we select 3k images from CSS \cite{vo2019composing} Dataset, which should have at least 4 objects in the scenes. Then we add material information to these objects, which means randomly choosing the metal or rubber for the object. According to the query answer (generation process is explained in section \ref{subsection: Query Generation}), we modify the scene files in CSS. Then following the image generation engine provided by Liu \cite{liu2019clevr} \textit{et al.}, we create our own image database, which contains 26k images in the training set and 26k images in the testing set. The image database is larger than the previous CSS image database, we show details in Figure \ref{fig: CRIR Dataset}a. Our objects contain 4-dimension information, color, shape, size, material, comparing to 3-dimension in CSS.

\subsection{Query Generation}
\label{subsection: Query Generation}
\par The queries' generation can be summarized as follows:
\par \textbf{1.} We add 3 types of programs, {\tt add}, {\tt remove} and {\tt make}, to the CLEVR-programs universe. {\tt Make} programs are combined with a value chosen from 4 attributes, shape, color, material and size. For instance, in the query text ``make yellow sphere in front of green cube small", the program is ``make[small]". 
\par \textbf{2.} We inherit the CLEVR templates and throw away the templates that are unsuitable for retrieval queries. Specifically,  we select 5 templates from CLEVR\cite{johnson2017clevr}: \textit{Zere Hop}, \textit{One Hop}, \textit{Two Hop}, \textit{Three Hop}, \textit{Single And}, which are suitable for generating our image retrieval text queries.
\par \textbf{3.} We modify the CLEVR question generation engine, whose original output is the answer in CLEVR answer set. To be specific, we force our engine to output the objects' indices which need to be modified with the text query suitable for this task, unlike the questions in CLEVR. Firstly, we generate {\tt remove}, {\tt make} queries and maintain the number of {\tt remove} queries 2 times more than {\tt make} queries. Then we use these queries to change the image scenes according to the {\tt remove} or {\tt make} type and objects' indices. After that, we apply these scenes to render new images. For {\tt add} query generation, we compare the original scene with the generated scene one by one, if it reduces one object, we put it to an addition candidate list. After that, we randomly choose pairs from the list then change query type from {\tt add} to {\tt remove} and convert the $I_{target}$ to $I_{input}$ until the number of {\tt add} query text is equal to the {\tt remove} query text. Finally, we obtain 22.5k queries in training set and 22.5k queries in testing set.

\par \textbf{Results.} We compare CRIR dataset with CSS dataset which shows in Figure \ref{fig: CRIR Dataset}. To conclude, CRIR's image database is larger than CSS,  Figure \ref{fig: CRIR Dataset}a. CRIR's query text is more versatile than CSS's, Figure \ref{fig: CRIR Dataset}b. Additionally, CRIR's scale in query texts is larger than CSS's, Figure \ref{fig: CRIR Dataset}c. In Figure \ref{fig: CRIR Dataset}d, we show the details about the types of query text in CRIR. All these perspectives above demonstrate we generate a more complex reasoning and hard-to-edit dataset for text-editing image retrieval task.
\section{Experiment}
\label{section-5}

\par To display the effectiveness of our model on editing the scene graphs, we carry out experiments on CSS and CRIR. Our measurement is set to recall at rank 1 in order to better compare with other methods.
\subsection{Experiment on CSS Dataset}
\label{section: experiment on css}

\par Because nearly 85\% percent query texts in CSS dataset has exact location word, we create exact location program for this dataset, such as ``Location[B-L]'' program in Figure \ref{fig: CSS reasoning}, which can attend the object at Bottom-Left grid. The location boundaries used to split the 3 by 3 grids are (-0.99, 0.86) horizontally and (-0.47, 2.35) vertically. In CSS training, we randomly select 30 text queries (add, remove, make, each 10) and annotate them for pretraining our program generator, with learning rate $6 \times 10^{-4}$, $12000$ iterations, and batch size 64. Especially, in CSS, we keep the structure of Scene Graph when ME executing programs to modify it. We show an example of ME's reasoning details in Figure \ref{fig: CSS reasoning}.

\begin{table}[t]
\begin{center}
\begin{tabular}{ccccccccc}
\hline
Method  & ImageO & TextO & Concate & MRN\cite{MRN}  & Relat\cite{Relationship} & FiLM\cite{perez2018film} & TIRG \cite{vo2019composing} & Ours \\ \hline
Add     & 0.1    & 0.1   & 73.6    & 72.9 & 74.7  & 75.9 & 83.6 & \textbf{99.7} \\
Remove  & 0.2    & 0.1   & 45.3    & 42.8 & 49.8  & 52.3 & 64.3 & \textbf{99.8} \\
Make    & 9.1    & 0.1   & 63.2    & 64.6 & 61.8  & 68.6 & 73.3 & \textbf{99.9} \\
Overall & 6.3    & 0.1   & 60.6    & 60.1 & 62.1  & 65.6 & 73.7 & \textbf{99.8} \\ \hline
\end{tabular}
\end{center}
\caption{Quantitative results in CSS. The retrieval metric is Recall at rank 1. ImageO and TextO means Image Only and Text Only, respectively. Concate is a method just concatenating language feature and image feature without other processing. As show above, our model with GED reward surpasses all other models by large margins in all three types of queries.}
\label{tab: CSS}
\end{table}

\par In Modifying Engine, we inherit programs which have prefix {\tt filter} from CLEVR and create corresponding modules for them. Here we list some special programs outside CLEVR-programs universe and its corresponding modules' operation details, which shows as follows:

\par \textbf{Location Modules.} Location modules are instantiated from location programs and they will select the exact location like ``Top-Left'' for operating modification later on. There are totally 9 different location modules corresponding to 3 by 3 grids. 

\par \textbf{Remove, Add, Make Modules.} According to the start word of the query text we can obtain {\tt Remove}, {\tt Add}, or {\tt Make} programs to instantiate modules and execute operations over structural scene graph. All these three types of modules will be executed last. We also create 2 reasoning modes, ``normal'' and ``add''. Mode ``normal'' is created for {\tt Remove} and {\tt Make} modules while mode ``add'' for {\tt Add} module. {\tt Remove} module is applied to change every attribute value of the attended nodes into {\tt NULL} while {\tt Make} modules change the specific attribute that attended by the previous modules. But in {\tt Add} type queries, some attributes will not be assigned value, for example, in query text ``Add small cube to the top-left'', the color information is not assigned. Thus we create value 1 for these unassigned attributes when modifying the node in Add. In Graph Edit Distance \cite{sanfeliu1983distance} computation, value 1 can match any value except {\tt NULL}.

\textbf{Graph Edit Distance in CSS.} In the CSS settings, scene graphs are structural and each graph has 9 nodes with rigid location. Thus in equation \ref{ecgm}, only substituting cost exists. The Graph Edit Distance \cite{sanfeliu1983distance} computation can be simplified as:
\begin{equation}
    GED = \min\big(\sum_{x \in \hat{V_1}}{c_{ns}(x)}\big).
\end{equation}
For every node having 3 attributes, we set the cost of substituting every attribute to $1 / 3$ and force the reward to be 0 if GED between 2 graphs is larger than 1. In the testing, we apply graph edit distance as our retrieval metric. If GED of two pairs' graphs happen to be the same, we will randomly choose one as our target image. We return this reward to finetune the program generator with 50,000 iterations and early-stopping strategy. Batch size is fixed to 64 in finetuning.

\textbf{Results and Analysis.} As table \ref{tab: CSS} shows, our model achieves state-of-the-art result and surpass the previous baseline model TIRG \cite{vo2019composing} drastically no matter in which query text type. Additionally, We give an example in CSS and display our symbolic model's transparent and powerful reasoning ability in Figure \ref{fig: CSS reasoning}. The results and the visualization prove our method's effectiveness on editing symbolic scene graph according to the instructions.

\par One biggest problem in CSS is that if our model predicts the correct location program, it will attend to the correct object because of the exact location information text provided. Besides, all queries in CSS are \textit{Zero Hop}, which does not need any edge information for reasoning. Only through attribute {\tt filter} or location programs can the model attend to correct nodes. Therefore, we propose a more complex dataset to validate our model's reasoning ability: \textbf{1.} To simulate the real-world scene, the dataset should just provide the relational location information, not exact locations. \textbf{2.} The dataset should also contain more complex queries, not just \textit{Zero Hop}. Aiming at this, we generate CRIR. Details can be referred to Section \ref{subsection: Query Generation}.


\begin{table}[t]
\begin{center}
\begin{tabular}{ccccccc}
\hline
Model       & Zero\_Hop & One\_Hop & Two\_Hop & Three\_Hop & Single\_And & Overall \\ \hline
Concatenate & 45.1      & 25.2     & 16.7     & 3.7        & 15.8        & 21.3    \\
MRN\cite{MRN}& 44.8     & 28.3     & 19.4     & 3.5        & 19.6        & 23.1    \\
Relat\cite{Relationship} & 51.9& 28.2& 21.3   & 4.0        & 18.1        & 24.7    \\
FiLM\cite{perez2018film} & 50.2& 28.1& 21.1   & 3.6        & 16.3       & 23.9 \\
TIRG\cite{vo2019composing} & 53.8      & 29.6     & 22.9     & 4.8        & 19.4        & 26.1    \\
0,1 Reward  & 95.9      & 95.4     & 95.8     & 95.3       & 94.5        & 95.4    \\
GED Reward  & \textbf{99.0} & \textbf{98.4} & \textbf{98.2} & \textbf{97.9}       & \textbf{98.4}        & \textbf{98.4}    \\ \hline
\end{tabular}
\end{center}
\caption{Quantitative results in CRIR. The retrieval metric is Recall at rank 1. We adapt Vo's \cite{vo2019composing} code to train TIRG and Concat models in CRIR. 0, 1 means applying 0, 1 reward to finetune our model. GED means applying GED reward to optimize our model. In particular, our model with GED reward optimizing surpass all the other models in every type of query.}
\label{table: CRIR experiment result table}
\end{table}

\subsection{Experiment on CRIR Dataset}
\label{section: experiment on CRIR}
\par In CRIR dataset, we remove the location programs to simulate a more realistic situation. Though we can still generate the structural scene graph shown as Figure \ref{fig: CSS reasoning}, we consider that 3 by 3 grid structural scene graph is a special case in the real world. Therefore, we cancel this rigid location and generate relational scene graphs for the images in the CRIR dataset then editing relational scene graphs. In training procedure, we randomly select 300 programs (add, remove, make type each 100 programs) to pretrain our program generator with 16,000 iterations and $7 \times 10^{-4}$ learning rate. {\tt Add}, {\tt Remove} modules in CRIR settings are different from CSS. To illustrate, {\tt Add} module in CRIR Modifying Engine will add a pseudo node to the scene graph, unlike {\tt Add} module in CSS just changing the attended node. We give an instance here. Assuming the query text is ``Add red large sphere to the left of large yellow cylinder", our Modifying Engine will execute {\tt Add} module to create one pseudo node. The pseudo node will create edges linking to all other nodes in scene graphs. Then if PG outputs ``Relate[left]'' program, it will assign edge linking to the ``large yellow cylinder'' {\tt Left} value. For {\tt Remove} module, unlike the example shows in Figure \ref{fig: CSS reasoning} just changing the attributes to {\tt NULL}, in CRIR, {\tt Remove} module will remove the attended objects and all the edges of the nodes as example shows in Figure \ref{fig: CRIR Reasoning}.

\textbf{Graph Edit Distance in CRIR.} We apply graph edit distance to finetune our program generator. Unlike the graph edit distance in CSS dataset without edge information, in this part, our graph edit distance has all costs (four parts) in equation \ref{ecgm}. And we set the cost of inserting a node to 1, the cost of deleting to 1, substituting an attribute value to $1 /4$, and substituting an edge to $1/16$. We only return the positive reward and force negative reward to 0. We return this reward to finetune PG with 60, 000 iterations, early-stopping training strategy and batch size is fixed to 64.

\textbf{Results and Analysis.} We provide the quantitative results in table \ref{table: CRIR experiment result table}. As it shows, our model can still maintain high-level performance in all five types of queries while TIRG\cite{vo2019composing} and Concat models drop its performance on CRIR drastically. \textit{Zero Hop} is the easiest type query text so all the model achieve its best performance on this type of query text. From table \ref{table: CRIR experiment result table}, we also discover that \textit{Three Hop} is so complicated that TIRG and Concat model can not learn this type of query text well, but our models are able to comprehend this type of queries. Notably, Our GED reward is more efficient than simple 0, 1 reward according to the last 2 columns of table \ref{table: CRIR experiment result table}. The performance of our model with GED reward surpasses all other models in all types of queries, which proves the GED reward is a precise measurement of the final modified graph to the target graph. We also show one case of our model's (optimized by GED reward) symbolic reasoning insight in Figure \ref{fig: CRIR Reasoning}.

\par
\begin{figure}[t]
\begin{center}
\includegraphics[width=0.8\linewidth]{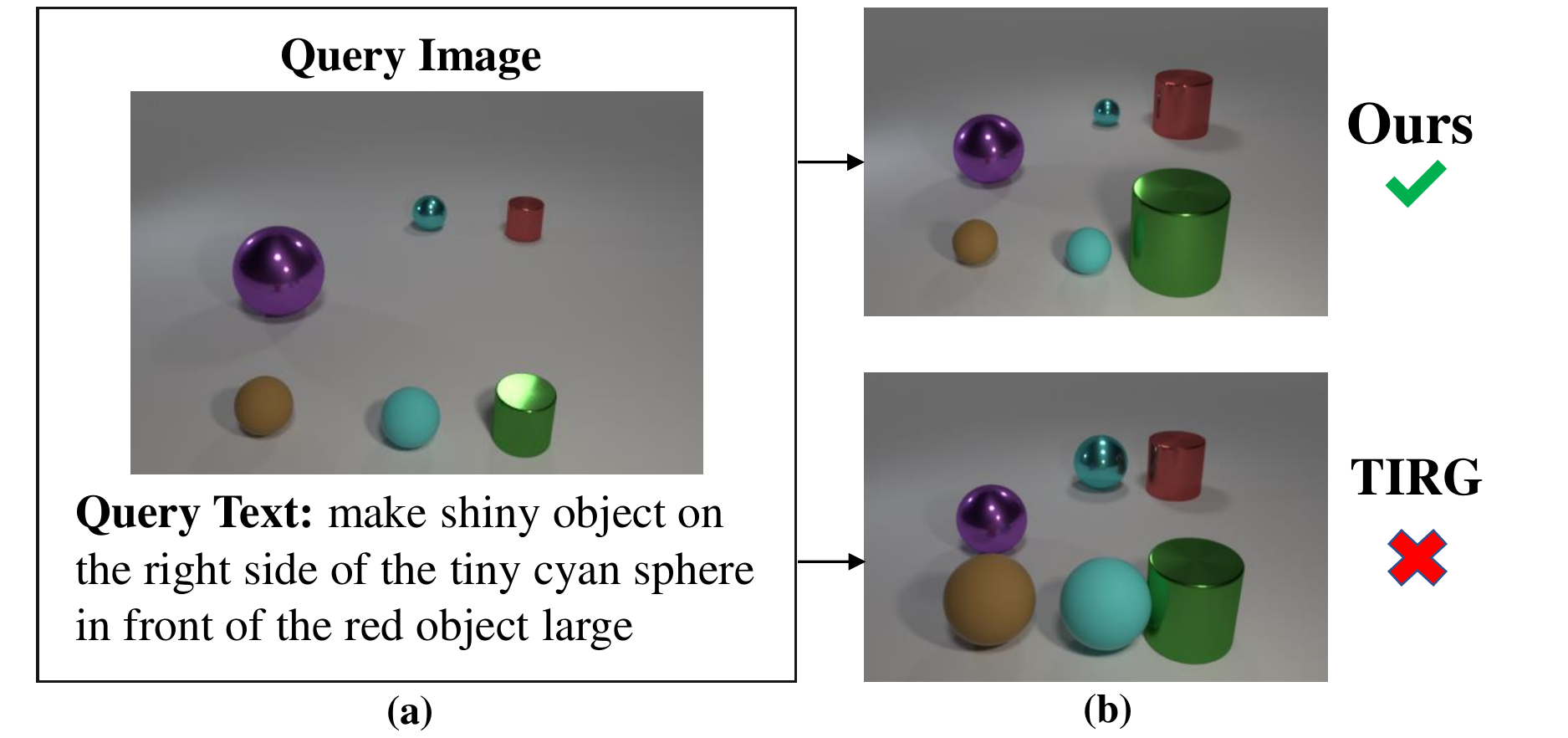}
\end{center}
  \caption{The qualitative results in CRIR. (a) represents the input query while (b) is the retrieval results by applying our model (with GED reward optimizing) and TIRG \cite{vo2019composing}  }
\label{fig: crir qualitative result}
\end{figure}

 We compare the performance of feature combination model and ours qualitatively and show a case in Figure \ref{fig: crir qualitative result}. As case shows, the TIRG\cite{vo2019composing} just retrieve the image with similar objects while our scene-graph-editing based model can understand the true meaning of this text and retrieve the correct image in CRIR dataset. In conclusion, our model has more powerful generalization ability than TIRG\cite{vo2019composing}.

\section{Conclusion}
\label{section-7}

\par In this paper, we have proposed a GED reward for learning to edit scene graph in the context of text-editing image retrieval. By exploiting CLEVR toolkit, we generate a Complex Reasoning Image Retrieval dataset, simulating a harder case of editing. To the best of our knowledge, we are the first to apply Graph Edit Distance as retrieval metric. Furthermore, we validate our model on both CSS and CRIR dataset and achieve nearly perfect results, surpassing other models by large margins, proving the effectiveness of editing scene graphs in the context of text-editing image retrieval. However, our model still need prior information like the annotations of some latent programs to pretrain the PG. We expect our machine can own the reasoning ability without using any prior information in the future. 
\\
\textbf{Acknowledgments.} This research was supported by the National Research Foundation Singapore under its AI Singapore Programme (Award Number: AISG-RP-2018-003) and the MOE Tier-1 research grants: RG28/18 (S) and RG22/19 (S). Q. Wu's participation was supported by NSFC 61876208, Key-Area Research and Development Program of Guangdong 2018B010108002.

\clearpage

\bibliographystyle{splncs04}
\bibliography{egbib}
\end{document}